\def\year{2020}\relax
\documentclass[letterpaper]{article} % DO NOT CHANGE THIS
\usepackage{aaai20}  % DO NOT CHANGE THIS
\usepackage{times}  % DO NOT CHANGE THIS
\usepackage{helvet} % DO NOT CHANGE THIS
\usepackage{courier}  % DO NOT CHANGE THIS
\usepackage[hyphens]{url}  % DO NOT CHANGE THIS
\usepackage{amsmath}
\usepackage{amssymb}
\usepackage{bm}
\usepackage{multirow}
\usepackage{graphicx} % DO NOT CHANGE THIS
\usepackage{xcolor}
\usepackage[titlenumbered,ruled]{algorithm2e}
\usepackage{graphicx}  % DO NOT CHANGE THIS
\newcommand{\somakcomment}[1]{\textcolor{red}{[Somak:#1]}}

\begin{document}

\title{Uncovering Relations for Marketing Knowledge Representations}

\author{Somak Aditya, Atanu R Sinha\\
Adobe Research\\
{\tt\small \{saditya,atr\}@adobe.com}
}

\maketitle

\begin{abstract}
\iffalse
\somakcomment{Make crisp and shorten: Marketing entails consumers' and firms' behaviors, while generating lot of data. These data in turn form the basis of analysis, interpretation, insights, explanations, all leading to myriad marketing decisions by firms. While data and models grow in sophistication, decision making in many instances is not solely an outcome of those. Human managers bring to bear marketing knowledge which reside outside of data and models to make decisions. It behooves creation of an automated marketing knowledge base with the goal of informing decisions making as well as interacting with data and models.} 
\fi

Online behaviors of consumers and marketers generate massive marketing data, which ever more sophisticated models attempt to turn into insights and aid decisions by marketers. Yet, in making decisions human managers bring to bear marketing knowledge which reside outside of data and models. Thus, it behooves creation of an automated marketing knowledge base that can interact with data and models. Currently, marketing knowledge is dispersed in large corpora, but no definitive knowledge base for marketing exists. Out of the two broad aspects of marketing knowledge - representation and reasoning - this treatise focuses on the former. Specifically, we focus on creation of marketing knowledge graph from corpora, which requires identification of entities and relations. The relation identification task is particularly challenging in marketing, because of the non-factoid nature of much marketing knowledge, and the difficulty of forming rules that govern relations. Specifically, we define a set of relations to capture marketing knowledge, propose a pipeline for creating the knowledge graph from text and propose a rule-guided semi-supervised relation prediction algorithm to extract relations between marketing entities from sentences. %Our work makes a contribution in relation prediction in marketing knowledge. 
%[Somak, to be refined and more to be added]      ]
\end{abstract}

\noindent 
\section{Introduction}
Effective decision making to choose marketing actions is much more than utilization of data, reporting tools and models offered by today's advanced Analytics capabilities. Decision making is part art, part science\footnote{ (\url{https://hbswk.hbs.edu/item/making-right-choices-art-or-science})}. While the ``science'' of marketing decision making captures research imagination and offers great advances, the ``art'' of marketing decision making lags behind. 
%\somakcomment{How do we know its unstructured knowledge that humans use}. 
The art includes 
%\textcolor{red}{unstructured knowledge} 
knowledge humans use to overlay on structured information from data, tools, and models, to make decisions and choices. Part of the knowledge resides inside humans, and others lie in corpora of text books, business articles, experts' writings, research papers, and case studies. With focus on the latter, our objective is to give this knowledge shape in the manner of a Marketing Knowledge Representation (MKR) for it to interact with structured information from data, tools, and models, to eventually advance decision making. The paper exposes the specific challenges of creating MKR, relative to other forms of KR, and then addresses some of them. 

For concreteness, consider segmentation, a regularly occurring, fundamental task in marketing decision making. Data from transaction and clickstream capture consumer behavior and are added to data on marketing actions and consumer demographics. Models attempt to estimate the differential effects of marketing actions across consumers on business-desired outcomes and map demographics to those effects to divide the consumers into different segments. When presented with these results, a seasoned human marketer's knowledge suggests that for effective segmentation she needs to look beyond demographics, into other characteristics, say, psychographics. 
%\somakcomment{Its a bit heavy claim. Maybe soften it as we are not doing the reasoning.}
%\textcolor{red}{To replace or minimize the seasoned manager's role in the segmentation task, we seek an automated system which encodes this knowledge and offers advice to the decision making task when presented with results from data and models}. 
In this paper, we demonstrate an approach to encoding such knowledge into an MKR. Once an MKR is built, the next step involves building a reasoning engine on the MKR to move toward automated decisions. Staying within segmentation, a reasoning engine can explain whether psychographics segmentation is the way to go, given findings from data and models. In this paper we focus on knowledge representation, but not on reasoning.

Our representation takes the form of a knowledge graph (KG), where the graph ``mainly describes real world entities and their interrelations'' \cite{paulheim2017knowledge}. %``A knowledge graph mainly describes real world entities and their interrelations, organized in a graph \cite{paulheim2017knowledge}''. 
The objectives we pursue are organization of marketing information, non-factoid concepts and results from marketing academic literature in a Marketing domain specific KG (MKG). A KG embodies nodes and edges, where nodes are {subject, object} and edges are {relations}. The problem of extracting triples, defined as $\langle$subject, relation, object$\rangle$, from Marketing corpora is challenging for multiple reasons: (1) much of marketing knowledge is non-factoid; (2) entities do not have a taxonomy; (3) the typical corpora is not tightly worded leading to non-informative content; (4) entities are longer sequence of words; (5) relations are marketing domain specific and cannot be necessarily drawn from existing sources of relations such as ConceptNet; (6) supervised approaches for relation prediction cannot be used due to severe labeling limitations. Specifically, the current effort addresses the challenge of predicting relations using a semi-supervised approach and based on a relatively small set of labeled relations. The experiments in this work are based on our efforts of creating an MKG from the chapter on Segmentation in a marketing textbook.   

Our main contributions in relation prediction are: (1) demonstrating an approach in creating a Marketing Knowledge Graph, with (2) semi-supervised Relation prediction, using (3) Rule-regularization, given relatively few labeled relations.

\iffalse
Marketing decision makers and marketing analysts alike bring their ``external'' knowledge of marketing that is, external to data, reports and models - to generate insights and explanations that inform decisions. We offer an approach that helps creation of knowledge graph (KG), from which such knowledge can be drawn automatically. In defining KG, we borrow from [1] ``A knowledge graph mainly describes real world entities and their interrelations, organized in a graph.`` To serve our purpose of marketing analysis and decision making, the objectives we pursue are organization of marketing information, non-factoid concepts and results from marketing academic and practitioner literatures in a KG. By ingesting received wisdom available in various corpora, a marketing domain KG can interact with models and analysis run on data to improve insight generation and explainability. 
The problem we address is - how to create a marketing domain knowledge graph from any marketing corpora?
\fi

%[1] https://hbswk.hbs.edu/item/making-right-choices-art-or-science

%[2] Paulheim, H. (2017). Knowledge graph refinement: A survey of approaches and evaluation methods. Semantic web, 8(3), 489-508.

\iffalse
\begin{itemize}
\item You must use the 2020 AAAI Press \LaTeX{} style file and the aaai.bst bibliography style file, which are located in the 2020 AAAI Author Kit (aaai20.sty and aaai.bst).
\end{itemize}
\fi

\section{Related Literature}

Our work is closely related to efforts in commonsense knowledge representation, automatic knowledge base construction, relation extraction from text and knowledge integration in deep neural networks. 

\textbf{Commonsense Knowledge Representation}: Proposing task-independent knowledge representation for any domain has been a central challenge for the KR\&R community. Many commonsense KGs that capture ontological, causal, and other types of common-sense relations between general-domain concepts have been fairly popular such as ConceptNet \cite{speer2017conceptnet}, Cyc \cite{Lenat:1990:CTP:79173.79176} and WordNet \cite{miller1995wordnet}. Domain-specific knowledge bases such as AURA-KB \cite{Barker:2007:AES:1619797.1619980} built on top of the Knowledge-machine ontology for encoding knowledge in biology books have seen some adoption. For marketing domain, the semantics of general-world relations and concepts become ambiguous. Also, our search did not produce any KG specifically for marketing. Our experiments show that existing KGs do a poor job of representing knowledge in Marketing domain, due to the nuanced and non-factoid nature of knowledge in this domain, and the emphasis of these KGs on representing general world knowledge. This makes our effort necessary.  %Herein lies one of our major contributions in the sense, that entities in marketing do not follow a well-defined ontology and often all knowledge base construction techniques depend on or benefit from the entity type information.  

\textbf{Automatic Knowledge Base Construction}:
KGs have been traditionally constructed using curated (WordNet), semi-curated (ConceptNet \cite{speer2017conceptnet}), fully automated (YAGO, NELL) approaches. Curated approaches pose very costly for marketing domain. Hence, automated knowledge base construction or completion cannot be avoided. KB construction has made strides with the use of knowledge graph embedding \cite{wang2017knowledge}. The continuous vector representation in low dimensions allows capturing latent semantic relations and applying vector algebra for inferencing about relations. In turn, this affords flexibility for tasks ranging from relation prediction, to entity resolution, to knowledge graph completion. One class of methods perform the embedding task by matching embedding to facts available on the knowledge graph. Other class of approaches uses additional information that are available \cite{wang2017knowledge}. This information includes types of entity, description and logical rules. The embeddings consider either instances of real-world entities in the knowledge graph, or, ontological concepts of the knowledge graph, but not both. More recent work \cite{hao2019universal} advances representational learning by capturing knowledge jointly in both real-world entities and in ontological concepts, as well as, in links that connects them. With focus on relation prediction, our work follows in this tradition of using knowledge graph embeddings.

\textbf{Relation Extraction}: For the task of relation extraction from text, the relation between two concepts or entity mentions in a sentence is mapped to one of the classes in a predetermined closed set of relations. The relevant literature on methods can be grouped as: i) rule-based, ii) supervised and semi-supervised, iii) link prediction. Research in relation extraction has moved from applying hand-coded rules to extract relations \cite{rosemblat2013methodology}, to using hand-engineered features and strong classifiers \cite{kambhatla2004combining,minard2011hybrid} to classify relations between entities. However, given the brittleness of manually designed rules or features, and availability of large amount of data, the focus has shifted to different end-to-end neural models such as convolutional neural networks \cite{zeng2014relation}, recursive neural network \cite{ebrahimi2015chain}, and long short-term memory network \cite{miwa2016end}. %in \cite{rosemblat2013methodology} applies hand-coded rules to extract relations between bio-medical entities. While the recall of such tools were above 80$\%$, precision was comparatively lower. Also rule-based systems overall has been proven to be brittle. Among supervised methods, the machine learning community has experimented with both hand-engineered feature based \cite{kambhatla2004combining,minard2011hybrid} and end-to-end deep learning methods.
%Provided the limitations of hand-engineered feature-based supervised classification methods , the community has moved to deep learning based end-to-end relation extraction techniques.
Work in link prediction \cite{ostapuk2019activelink} has also inspired use of information from available knowledge graph for relation prediction tasks \cite{xu-barbosa-2019-connecting}. 
One obstacle in employing successful supervised classifiers is the dearth of large human-annotated data set of labels. Hence, semi-supervised approaches are receiving attention. Some work model this problem as a multi-instance learning problem \cite{riedel2010modeling}, and improve the overall accuracy through distant supervision and active learning \cite{sterckx2014using}. Under distant supervision, the problem of predicting relations from noisy annotations is tackled by \cite{feng2018reinforcement} using reinforcement learning. A recent paper \cite{lin2019learning} takes an important step forward by jointly optimizing the dual tasks of retrieving sentences given a relation and predicting a relation in a given sentence (hereafter, DualRE). Rather than self-selection, both prediction and retrieval module annotate unlabeled sentences and provide data to each other, thus potentially curbing the limited supervision issue. %``Specifically, as the prediction and retrieval models annotate/retrieve unlabeled sentences from the auxiliary corpus, both models provide additional training data to each other and thus help resolve the insufficient supervision issue \cite{lin2019learning} pp. 2.`` 

Annotations of relations for sentences in Marketing corpora are generally not available. There is need for marketing expertise to annotate relations in order to obtain high quality labels. Relatively few labels can be annotated and that too at significant cost in time and money. Given the unusually low labels, we look towards encoding knowledge using rules that govern the relations and take inspiration from the knowledge integration work in deep neural networks \cite{hu-etal-2016-harnessing,guo2018knowledge,wang2019logic}. However, to the best of our knowledge, these work do not integrate weighted First Order Logic rules in a semi-supervised scenario. Given our goal of relation prediction in  marketing corpus and faced with a small set of labeled relations and a large set of unlabeled corpus, we improve upon the DualRE approach by integrating knowledge from weighted logical rules.

\section{Background: Markov Logic Network}
   Markov Logic Network (MLN) \cite{richardson2006markov} is a popular probabilistic logical framework that uses weighted First Order Logical (FOL) formulas to encode an undirected, grounded  probabilistic graphical model (i.e. Markov Network).  %The Markov Logic Network (MLN) is targeted to use the full expressiveness of First Order Logic. 
   The rules in MLN are weighted so that the strict constraints of hard rules (rules that are satisfied always) are eliminated to model the real world more efficiently. It retains the flexibility of modeling hard FOL rules by adding hard constraints as well. 
   Formally, an MLN $L$ is a set of pairs $\langle F,w \rangle$, where $F$ is a first order formula and $w$ is either a real number or a symbol $\alpha$ denoting hard weight. Together with a finite set of constants $C$, a Markov Network $M_{L,C}$ is defined as containing: i) one binary node for each grounding of each predicate appearing in $L$; and ii) one feature for each grounding of each formula $F_i$ in $L$. The value of feature is 1, if grounded formula is true; 0, otherwise.
   The probability distribution over possible worlds $x$ specified by the ground Markov Network $M_{L,C}$ is given by:
   \begin{equation*}
   P(X=x) = \frac{1}{Z} \exp (\sum_{i=1}^F w_i n_i(x))
   \end{equation*}
   where $F$ is the number of grounded formulas, $n_i(x)$ is the number of true groundings of the formula $F_i$ in the world $x$. The MLN inference is equivalent to finding the maximum probable world according to the above probability formulation. Weight learning is done by maximizing the pseudo-likelihood.
   
\section{Marketing Knowledge Representation}

\begin{table}[!htpb]
\resizebox{\columnwidth}{!}{%
\begin{tabular}{|l|l|}
\hline
\multicolumn{1}{|c|}{\textbf{Sentence}}                         & \multicolumn{1}{c|}{\textbf{Triplets}}                                                                        \\ \hline
\multirow{4}{*}{\begin{tabular}[c]{@{}l@{}}If preferences are relatively \\ homogenous within a segment, \\ the positions of competing \\ brands will be relatively similar, \\ and the quantity of advertising \\ and promotion will be critical \\ competitive weapons.\end{tabular}} & \begin{tabular}[c]{@{}l@{}}HasProperty(segment, \\ homogeneous preferences\end{tabular}            \\ \cline{2-2} & \begin{tabular}[c]{@{}l@{}}LeadsTo(competing brands, \\ positions)\end{tabular}                            \\ \cline{2-2} 
 & \begin{tabular}[c]{@{}l@{}}HasProperty(positions, \\ similar)\end{tabular}          \\ \cline{2-2} 
 & \begin{tabular}[c]{@{}l@{}}HasProperty(competitive weapons, \\ quantity of advertising and promotion)\end{tabular}            \\ \hline
\multirow{2}{*}{\begin{tabular}[c]{@{}l@{}}Segments often overlap, making it \\ difficult  to position products in \\ different segments independently.\end{tabular}}                                                                                                             & \begin{tabular}[c]{@{}l@{}}ObstructedBy(position product, \\ segments often overlap)\end{tabular} \\ \cline{2-2} 
      & \begin{tabular}[c]{@{}l@{}}MotivatedByGoal(position products,\\ different segments)\end{tabular}      \\ \hline
\multirow{3}{*}{\begin{tabular}[c]{@{}l@{}}We must balance the costs \\ of positioning with price and \\ share changes to identify the\\  strategy that will achieve \\ maximum long-run profitability.\end{tabular}}                                                                   & \begin{tabular}[c]{@{}l@{}}DependsOn(costs of positioning, \\ price)\end{tabular}                      \\ \cline{2-2} 
& \begin{tabular}[c]{@{}l@{}}DependsOn(costs of positioning, \\ share changes)\end{tabular}                      \\ \cline{2-2} 
& \begin{tabular}[c]{@{}l@{}}MotivatedByGoal(strategy, \\ maximum long-run profitability)\end{tabular}          \\ \hline
\end{tabular}%
}
\caption{Triplets from illustrative sentences}
\label{tab:sentence_triplet}
\end{table}
%\end{table}

%\somakcomment{Can you please shorten this - and may be just remove one example.}
We first note the idiosyncrasies of marketing corpora to argue that (i) semantics of marketing-concepts do not map to common notion of entities, and (ii) relations in marketing are not adequately captured in sources such as ConceptNet5. Marketing-concepts are compound and much information is not commonsense knowledge. Consider the sentence on the vital topic of positioning. ``If we are to make good positioning decisions, we need to know what dimensions do consumers use to evaluate competitive marketing programs.'' For an MKR, the marketer relevant information in this sentence is a set of triplets: \textit{HasPrerequisite}(\textit{positioning decisions, know what dimensions}),  \textit{UsedBy}(\textit{know what dimensions, consumers}),  \textit{UsedFor}(\textit{know what dimensions, evaluate competitive marketing programs}). A few notables are: ``know what dimensions'' implicitly means ``knowledge of product dimensions''; those ``product dimensions'' that are \textit{UsedBy} ``consumers'' and \textit{UsedFor} ``evaluate competitive marketing programs;'' where ``evaluate'' is a short hand for ``evaluation.'' The last concept ``evaluate competitive marketing programs'' is an amalgam of three entities ``evaluate'', ``competitive'', and ``marketing programs''. Splitting into three entities explodes the set of nodes without adding to generality of representation. Moreover, entities do not form an ontolgy which can be exploited. Coming to relations in the above sentence, we use (\textit{HasPrerequisite}, \textit{UsedFor}) from ConceptNet5, and add a new relation, \textit{UsedBy}, as needed for marketing corpora. See Table \ref{tab:relations} for some new and ConceptNet5 relations used. Yet another sentence reads, ``Product positioning takes place within a target market segment and tells us how we can compete most effectively in that market segment.'' In essence the sentence states that ``product positioning is important to understand ``target market segment,'' and guides competition in the market segment. A KR shows; \textit{UsedFor}(\textit{product positioning, target market segment}) and \textit{RelatedTo}(\textit{target market segment, compete most effectively}). The use of \textit{RelatedTo} is not a precise association; however, as a form of general knowledge captures the essence of association. 

Consider another compound sentence, ``Segmentation analysis tells us how the market is defined and allows us to target one or more market opportunities.`` A KR takes the form of \textit{UsedFor}(\textit{segmentation analysis, how the market is defined}), and \textit{UsedFor}(\textit{segmentation analysis, market opportunities}). Since \textit{segmentation analysis} only makes sense within the context of a market we can add clarity, without losing any generality, by pre-fixing ``market.' Hence, it gives, \textit{UsedFor}(\textit{market segmentation analysis, how the market is defined}), and \textit{UsedFor}(\textit{market segmentation analysis, target market opportunities}). Note that \textit{how the market is defined} and \textit{target market opportunities} are key aspects of segmentation performed by marketers, and important concepts to be represented in a KR, along with their relations to segmentation analysis. Additionally, ``market segmentation analysis' is equivalent to ``market segment analysis' and a KR must recognize these similarities since both renditions appear in corpora. More examples of annotating triplets from sentences are found in Table \ref{tab:sentence_triplet}. Sentences in a marketing corpora are often written in an indirect style, making extrication with any existing parser prone to significant inaccuracies. The challenge is in devising a pipeline which can predict relations among these non-factoid, compound concepts; and as well, recognize when different variations of a concept mean the same thing. Importantly, we want relatively few relations which capture more general rules that govern association among different concepts. The complete set of relations, their semantics and examples used for our experiments are shown in Table \ref{tab:relations}.
%The relation prediction challenge is exacerbated by style of these sentences and our goal is to offer an approach which is agnostic to style or specific content. 

\section{Marketing Knowledge Acquisition Pipeline}

We adopt a pipeline-based approach, which has four stages: i) definition sentence extraction, ii) candidate triplets prediction, iii) relation extraction, and iv) merging.  A book chapter can be divided into definitions of important marketing terms and the rest of the content. The pipeline is described with respect to our example of the topic of Segmentation. 

1. Definition Sentences: For each definition of a marketing term such as ``segmentation'', we process them sentence by sentence.

2. Candidate Triplets: For each sentence, we parse using the Stanford syntactic dependency parser \cite{chen-manning-2014-fast} to get the syntactic parse tree and part-of-speech tags. We then use the parse tree (induced by syntactic dependency relations) and the part-of-speech tags to collect the set of all noun-phrases (NP), which do not include verb phrases or prepositional phrases. We treat each pair of NPs as a candidate for the next step. For example, ``Product positioning takes place
within a target market segment and tells us how we can compete most effectively in that market segment'', produces NPs ``product positioning'', ``target market segment'' and ``market segment''. 

3. Relation Extraction: For relation extraction, we pre-train a relation classifier which takes two noun-phrases, the sentence and positional part-of-speech and named entity tags. To train this classifier, we first consult a marketing expert to annotate correct relations for a small set of NP-pairs for the sentences from the textbook. Table \ref{tab:sentence_triplet} shows a few examples. We use this small labeled data and a large set of unlabeled data to train a semi-supervised relation classifier. This is a significant benefit of our approach.

4.  Merging: Using this classifier, we identify relations between all pairs of NPs from the previous step. This same classifier also informs which NP-pairs are not related by any relation. This step has two substeps: 4(a) To concentrate on the important entity1-relation-entity2 triplets; in the first sentence, we extract the list of NPs that are connected (via a path in the dependency graph) to the defined term, such as ``segmentation``.  This list becomes the next set of \textit{important} entities for the next sentence. We only concentrate on entities which are connected to the list of \textit{important} entities. 4(b) For the rest of the corpora, the hierarchical assumption over sentences is withdrawn. We extract entity1-relation-entity2 using similar method as in Steps 2 and 3. Given the graphs from 4(a) and 4(b), we merge using overlapping entities to arrive at the MKG.
%\end{enumerate}

 The complexity of recovering the interrelations between entities and mapping to a chosen set of well-defined relations are pushed to the relation extraction phase (Stage 3), which we describe next.

\section{Marketing Relation Prediction}

\begin{figure}
    \centering
    \includegraphics[width=8cm]{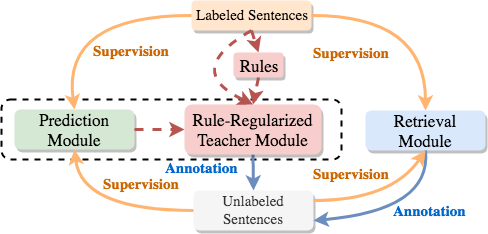}
    \centering
    \caption{Rule-regularized Selection in a Semi-Supervised Relation Prediction Framework}
    \label{fig:rulef}
\end{figure}

Relation prediction is the task of predicting a set of structured triplets (subject, relation, object) from a sentence encoding marketing knowledge. Figure 1 shows the framework. This process is performed in two steps: i) candidate relation mention extraction i.e. extracting $\mathbf{x} = \langle x, n_s, n_o\rangle$ from corpus where $x$ is a sentence, and $n_s$ and $n_o$ are marketing terms, ii) relation extraction, i.e. predicting a relation $r \in \mathcal{R}$ given a relation-mention $\mathbf{x}$.

\subsection{Relation-mention Extraction}
We get candidate NPs for each sentence $x$ from the second stage of our pipeline. %For each sentence $x$, we parse using the Stanford syntactic dependency parser \cite{chen-manning-2014-fast} to get the syntactic parse tree and part-of-speech tags. We then use the parse tree (induced by syntactic dependency relations) and the part-of-speech tags to collect the set of all noun-phrases (which does not include verb phrases or prepositional phrases). 
We heuristically eliminate NP-pairs that are connected via a path with length more than $P$ in the tree. This provides a set of unlabeled relation-mentions $U' = \{\textbf{x}_i\}_{i=1}^{N_{L+U}}$. We sample from this set and consult a marketing expert to provide correct labels for a small set of relation-mentions, which finally creates the set of labeled relation-mentions $L = \{\textbf{x}_i, y_i\}_{i=1}^{N_L}$ and set of unlabeled relation mentions $U = \{\textbf{x}_i\}_{i=1}^{N_U}$.

\subsection{DualRE: Semi-Supervised Relation Extraction}
%The relation extraction task (as defined in \cite{lin2019learning}) is to assign a relation label $r \in \mathcal{R}$ (a closed set of relations) to a relation-mention $\textbf{x} = \langle x, e_s, e_o \rangle$ where $x$ is a sentence, and $e_s$ and $e_o$ are subject and object entities ($e_s, e_o \in \mathcal{E}$). The entities in marketing knowledge are not restricted to a set of named entities. Hence, for our purposes, we represent the relation-mention by $\mathbf{x} = \langle x, n_s, n_o \rangle$, where $n_s$ and $n_o$ are elements of the set of marketing terms ($\mathcal{T}$).

Given a set of labeled  ($L$) relation-mentions and a set of unlabeled relation-mentions ($U$), our goal is to learn a relation prediction model $f$ that represents the training data $L$ and captures the information from the unlabeled data $U$. We follow the framework proposed in \cite{lin2019learning}. It consists of a prediction module $\mathcal{P}_\theta$ and a retrieval module $\mathcal{Q}_\phi$, where $\theta$ and $\phi$ are the model parameters. The prediction module's task is to represent the function $f$, i.e. predicting the relation $y$ given the relation-mention $\mathbf{x}$. It models the conditional probability $p_\theta(y|\mathbf{x})$ for a mention-label pair $(\mathbf{x},y)$. The retrieval module complements above by retrieving relevant relation-mentions given a specific relation. Hence, it models $q_\phi(\mathbf{x}|y)$ for a mention-label pair. As $q_\phi(\mathbf{x}|y) \propto q_\phi(\mathbf{x}, y)$ for a given relation $y$, the retrieval module estimates the joint probability and induces a ranking over different mentions $\mathbf{x}$ for a label $y$. The overall objective function is given by 
\begin{equation}
\begin{aligned}
    \mathbf{O} &= \mathbf{O}_P + \mathbf{O}_R +  \mathbf{O}_U,\\
    \mathbf{O}_P &= \mathrm{E}_{\mathbf{x},y \in L}[\log p_\theta(y|\mathbf{x})]\\
    \mathbf{O}_R &= \mathrm{E}_{\mathbf{x},y \in L}[\log q_\phi(\mathbf{x},y)]\\
    \mathbf{O}_U &= \mathrm{E}_{\mathbf{x} \in U}[\log p(\mathbf{x})]
\end{aligned}
\end{equation}
 $\mathbf{O}_P$ can be calculated using a cross-entropy loss between the ground truth and predicted labels, as shown in Equation 1. The objective $\mathbf{O}_R$ is approximated using a ranking loss:
\begin{equation}
    \mathrm{E}_{\mathbf{x},y \in L}[\log \sigma(\mathbf{z}^{T}\mathbf{y})] + \mathrm{E}_{\mathbf{x},y' \notin L}[\log (1-\sigma(\mathbf{z}^{T}\mathbf{y}'))],
\end{equation}
where $(\mathbf{x},y)$ is a labeled pair in $L$, $(\mathbf{x},y')$ is an incorrect relation pair with a relation mention $\mathbf{x}$, $\mathbf{z}$ is mention encoding for $\mathbf{x}$, $\mathbf{y}$ and $\mathbf{y}'$ are the embeddings of the relations $y$ and $y'$. 
  Lastly, $\mathbf{O}_U$ is approximated by the lower bound:
  $\mathbf{O}_U \geq \mathrm{E}_{\mathbf{x} \in U, y\backsim p_\theta(y|\mathbf{x})}[\log \frac{q_\phi(\mathbf{x},y)}{p_\theta(y|\mathbf{x})}]$. 
  
  \textbf{DualRE Learning Algorithm}: As proposed in \cite{lin2019learning}, an Expectation Maximization approach is used to jointly learn the modules. In the \textbf{E-step}, the prediction module $\mathcal{P}_\theta$ is learned by fixing $\mathcal{Q}_\phi$. Calculating the gradient of $\theta$ with respect to $\mathbf{O}$ amounts to:
  \begin{equation*}
  \begin{aligned}
      \nabla_\theta(\mathbf{O}) &= \mathrm{E}_{\mathbf{x},y \in L}[\nabla_\theta \log p_\theta(y|\mathbf{x})] \\&+ \mathrm{E}_{\mathbf{x} \in U, y\backsim q_\phi(y|\mathbf{x})}[\nabla_\theta \log p_\theta(y|\mathbf{x})],
\end{aligned}
  \end{equation*}
  where the first and second terms correspond to $\nabla_\theta(\mathbf{O}_P)$, and $\nabla_\theta(\mathbf{O}_U)$ respectively. Similarly, in the \textbf{M-step}, the retrieval module $\mathcal{Q}_\phi$ is updated fixing $\mathcal{P}_\theta$. The gradient with respect to $\phi$ is calculated as:
  \begin{equation*}
  \begin{aligned}
      \nabla_\phi(\mathbf{O}) &= \mathrm{E}_{\mathbf{x},y \in L}[\nabla_\theta \log q_\phi(\mathbf{x},y)] \\&+ \mathrm{E}_{\mathbf{x} \in U, y\backsim p_\theta(y|\mathbf{x})}[\nabla_\theta \log q_\phi(\mathbf{x},y)],
\end{aligned}
  \end{equation*}
  where the first and second terms correspond to $\nabla_\phi(\mathbf{O}_R)$, and $\nabla_\phi(\mathbf{O}_U)$ respectively. Both the steps require sampling from unannotated data. It is assumed that sampling from the averaged distributions, i.e. $p_\theta(y|\mathbf{x}) + q_\phi(y|\mathbf{x})$, is less noisy. Hence, samples are annotated using the intersection of these two modules before every iteration. For each iteration, the labeled dataset $L$ is added with the two modules' annotations (best predictions) to form $L_U$. Then $\mathcal{P}_\theta$ and $\mathcal{Q}_\phi$ are updated according to the E-step and M-step equations.

%   \textbf{Learning Framework}: We again follow the expectation-maximization approach (from \cite{lin2019dualre}) for jointly optimizing both modules. In the \textbf{E step}, we update the prediction module $\mathcal{P}_\theta$ by fixing the retrieval module $\mathcal{Q}_\phi$, which corresponds to the minimizing the KL divergence between $p_\theta$ and $q_\phi$. Whereas, in the \textbf{M Step}, we update the retrieval module $\mathcal{Q}_\phi$ by fixing the prediction module. The gradient calculations for parameters $\theta$ and $\phi$ are as given in the below equations:
%   \begin{equation}
%       \begin{aligned}
%       \nabla_\theta(\mathbf{O}) &= \mathrm{E}_{\mathbf{x},y \in L}[\nabla_\theta \log p_\theta(y|\mathbf{x})]  \\ &+ \mathrm{E}_{\mathbf{x} \in U, y\backsim p_\theta(y|\mathbf{x}) + q_\phi(\mathbf{x},y)]}[\nabla_\theta \log p_\theta(y|\mathbf{x})]\\ 
%     \nabla_\phi(\mathbf{O}) &= \mathrm{E}_{\mathbf{x},y \in L}[\nabla_\phi q_\phi(\mathbf{x},y)] \\&+ \mathrm{E}_{(\mathbf{x},y) \backsim p_\theta(\mathbf{x},y)+q_\phi(\mathbf{x},y)]}[\nabla_\phi \log p_\theta(y|\mathbf{x})]
%       \end{aligned}
%   \end{equation}
\begin{table*}[!htpb]
%\resizebox{\columnwidth}{!}{%
\begin{tabular}{|l|l|l|}
\hline
\multicolumn{1}{|c|}{\textbf{Wt.s}}                                          & \multicolumn{1}{c|}{\textbf{Rules}}                                                                                                                                                                                                                                                                                                     & \multicolumn{1}{c|}{\textbf{Semantics}}                                                                                                                                                                                                     \\ \hline
\begin{tabular}[c]{@{}l@{}}3.62\\ 3.63\\ 3.37\\ 0.48\\ 3.34\\ 1.43\end{tabular} & \begin{tabular}[c]{@{}l@{}}$enables(a1,a2) \leftarrow causes(a1,a2)$ \\ $\neg  hasLastSubevent(a1,a2) \lor \neg hasFirstSubevent(a1,a2)$\\ $ causes(a1,a2) \leftarrow affects(a1,a2)$ \\ $relatedTo(a1,a2) \leftarrow relatedTo(a2,a1)$ \\ $hasA(a1,a2) \leftarrow partOf(a2,a1)$ \\ $synonym(a1,a2) \leftarrow synonym(a2,a1)$\end{tabular} & \begin{tabular}[c]{@{}l@{}}\textit{causes} implies \textit{enables}.\\ a2 can not be both first and last sub-event.\\ \textit{affects} implies \textit{causes}.\\ \textit{relatedTo} is symmetric.\\ \textit{partOf} implies \textit{hasA}.\\ \textit{synonym} is symmetric.\end{tabular} \\ \hline

\end{tabular}%
%}
\caption{Set of rules used to act as constraint over the world of grounded predicates.}
\label{tab:rules}
\end{table*}

\begin{table}[!htpb]
\begin{tabular}{|l|l|}
\hline
\begin{tabular}[c]{@{}l@{}}1.0\\ 1.0\\ 1.0\\ 1.0\\ 1.0\\ 1.0\\ 1.0\\ 1.0\\ 1.0\\ 1.0\\ 1.0\\ 1.0\\ 1.0\\ 1.0\end{tabular} & \begin{tabular}[c]{@{}l@{}}dependsOn(Lifecycle, Trial)\\ dependsOn(Segmentation, Selection)\\ causes(Free\_samples, Potential\_purchasestate)\\ partOf(Product\_class, Brand)\\ dependsOn(Diffusion, Environmental\_change)\\ partOf(Home\_ownership, Religion)\\ hasProperty(Physical\_product, Distribution)\\ leadsTo(Product, Dimensions)\\ partOf(Sex, Demographic)\\ dependsOn(Trial, Sampling)\\ motivatedByGoal(Price\_reduction, Encroachment)\\ relatedTo(Place, Target\_market\_segment)\\ leadsTo(Product, Growth\_phase)\\ affects(Government\_regulations, Product\_lifecycle)\end{tabular} \\ \hline
\end{tabular}
\caption{Some examples of annotated ground-truth relations treated as predicates in Markov Logic Network.}
\label{tab:grounddb}
\end{table}

\subsection{Rule-Regularized Semi-supervised Relation Prediction}

Given the dearth of annotations in the marketing domain, we observe that prior rules over the relations can act as (global) constraints. A major drawback of the independence assumption of different samples is that the predictor ($\mathcal{P}_\theta$) is free to predict any conflicting relations between two concepts - such as A can not be both first and last sub-event of B (if B has more than one sub-event). Hence, weighted simplified rules can act as constraints. This requires us to solve two problems i) how to acquire the rules, and ii) how to integrate these rules with the predictor.

Firstly, these rules might be incomplete and should not be modeled as hard constraints. To model this ambiguity, probabilistic logical mechanisms such as MLN \cite{richardson2006markov} becomes a natural choice. The rules can be learned from the set of expert-provided ground truth relations using MLN's  standard structure learning algorithms \cite{kok2005learning}. In our case, the closed-world assumptions and sparse annotations force the MLN structure-learner to learn only unary clauses. Instead, we write the rules ourselves and then use MLN weight learning algorithm to learn the weights. We treat the ground-truth annotated relations as predicates of truth-value 1 (examples in Table \ref{tab:grounddb}), and use a few rules that can act as constraints. The rules and examples of ground truth are shown in Table \ref{tab:rules}. 
%\textcolor{red}{nothing wrong, a bit odd to use capital N as the index, small n as the cardinality}. 
Let the set of rules be denoted by $r_N(\mathbf{x},y)$, where $N={1, 2, \ldots, n}$. Using MLN's weight learning algorithm, we then learn the weights $\lambda_n$ for each rule in $r_N$. 

\begin{algorithm}[!htpb]
\SetAlgoLined
\SetKwInOut{Input}{input}
%\SetKwInOut{Output}{
\Input{Labeled data $L = \{\textbf{x}_i, y_i\}_{i=1}^{N_L}$, unlabeled data $U = \{\textbf{x}_i\}_{i=1}^{N_U}$, Weighted Rules $\{\lambda_N, r_N\}, N=\{1,\ldots,n\}$}
%\KwResult{Write here the result }
 Initialize: $L_U \leftarrow \emptyset$\;
 $p_\theta, q_\phi \leftarrow$ Pre-train prediction and retrieval module using L\;
 $t \leftarrow$ Compute from $p_\theta$ and $r_L(\mathbf{x},y)$ using $L$ (Eqn.~3))\;
 \While{$U \neq \emptyset$}{
  $L' \leftarrow$ Retrieve instances using intersection of $t$ and $q_\phi$ module\;
  Remove $L'$ from $U$ and add them to $L_U$\;
  Optimize $p_\theta$ using both $L$ and $L_U$ (Eqn.~1)\;
  Optimize $q_\phi$ using both $L$ and $L_U$ (Eqn.~2)\;
  Compute $t$ again using $L$ and $L_U$ (Eqn.~4);
 }
 \caption{Rule-Regularized DualRE Learning Algorithm}
 \label{algo:triple}
\end{algorithm}

\begin{table*}[!htb]
\begin{tabular}{|p{2.3cm}|p{7.9cm}|p{6.8cm}|}
\hline
Relations & Explanation & Example(s)\\
\hline
LeadsTo & A results in occurrence of B. The occurrence can be through other states, not necessarily direct. & Homogeneous preference (among consumers) (A) leads to competition (more competition) (B)\\
\hline
UsedBy & Usage of A by B for achieving some end state. Applies to both companies and consumers. & Product dimension (A) is used by consumer (B) to make a choice.\\
\hline
ImportantTo & A is a quality or characteristic that is salient to / for B. For a marketer, it is valuable to highlight some characteristics as particularly important, more than merely identifying them as a characteristic. & New dimension of product (A) is important to consumers (B). New dimension of product (A) is important for product positioning (B) by marketer.\\
%\hline
%DependsOn & A depends on B; however, is different from the inverse of LeadsTo. In marketing, A can depend upon several factors, B is among them. However, B may not lead to A. & Product positioning (A) depends on consumer evaluation of product attribute (B). Long run profit (A) depends on product positioning (B).\\
\hline
Affects & A can have impact on B, does not mean it will have an impact [inverse - AffectedBy]. & Government regulation (A) affects product life cycle (B). Style and fashion affects product life cycle. Political influence affects govenrment regulation.\\
\hline
Enables & A can facilitate the occurrence of B [inverse - EnabledBy]. & Good positioning is enabled by strong advertising claims. Perception and choice consumers form are enabled by product attribute.\\
\hline
% Below are ConceptNet5 relations with modified explanation &  &  \\
\hline
PartOf & A is a characteristic, which marketer associates with B. & Demographics (A) is a part of consumer (B). Price sensitivity (A) is a part of consumer preference (B).\\
% \hline
% HasPrerequisite & A is achieved by a marketer by performing B. & To determine competitive structure (A) a prerequisite is to understand preference difference [among consumers](B).\\
% \hline
% MotivatedByGoal & Goal of achieving B motivates marketer to perform A. & Price increase (A) is motivated by goal of pursuing higher profit (B).\\
\hline
HasFirstSubevent & A can start to happen when B starts to occur. & For diffusion of innovation (A) to occur the first subevent of adopt[ing] new product [by consumers](B) is necessary.\\
\hline
%CapableOf & Relative to ConceptNet5, CapableOf expanded from "Something that A can typically do is B" to Something that A can do is B, reducing focus on typicality. & Attitude [of consumers](A) is capable of identifying segmentation opportunities for marketer. Attitude is capable of differentiating behaviors [of consumers](B) by marketer.\\
\hline
HasA & A possesses certain traits B. A may not possess always. & Company (A) has strong patents (B). Consumer (A) has a higher price elasticity (B).\\
% \hline
% HasProperty & B is a property that characterises A; A possesses this property always. & Product (A) has the property that it requires distribution (B).\\
\hline
Synonym & A and B are often considered similar in what they convey. & Attitude segmentation (A) is synonymous with psychographics (B).\\
\hline
UsedFor & Purpose of A is to achieve B. & Perceptual map (A) is used by marketer to identify gaps (B) in marketplace.\\
%\hline
%CausesDesire & If a marketer experiences A, then it is likely the marketer wants to achieve B. & Shared production cost (A) among different products causes a desire to offer product line across segments (B).\\
\hline
RelatedTo & As in ConceptNet5, interpreted as a general relation. In marketing, many relations take this form, since pin pointing directionality is very difficult, without considering many other factors of context ad environment. & Maturity stage of product (A) in life cycle is related to product's ease of use by consumers (B).\\
\hline
Causes & A can cause B; although not always. In marketing, causal-relations are soft in scope, that is, does not mean A implies B [inverse - CausedBy]. & Good positioning (A) of a product causes high trial rate (B) of the product.\\ 
\hline
\end{tabular}
\caption{Illustrative relations with explanation and examples. The top five relations are new, while others come from ConceptNet5.  We also use \textit{CausesDesire, HasPrerequisite, MotivatedByGoal, HasProperty, DependsOn, CapableOf} are ommitted. These will be included in appendix.}
\label{tab:relations}
\end{table*}

\iffalse
\begin{table}
\centering
\begin{tabular}{|p{1.2cm}|p{3.5cm}|p{3cm}|}
\hline
Relations & Explanation & Example(s)\\
\hline
LeadsTo & A results in occurrence of B. The occurrence can be through other states, not necessarily direct. & Homogeneous preference (among consumers) (A) leads to competition (more competition) (B)\\
\hline
UsedBy & Usage of A by B for achieving some end state. Applies to both companies and consumers. & Product dimension (A) is used by consumer (B) to make a choice.\\
\hline
ImportantTo & A is a quality or characteristic that is salient to / for B. For a marketer, it is valuable to highlight some characteristics as particularly important, more than merely identifying them as a characteristic. & New dimension of product (A) is important to consumers (B). New dimension of product (A) is important for product positioning (B) by marketer.\\
\hline
DependsOn & A depends on B; however, is different from the inverse of LeadsTo. In marketing, A can depend upon several factors, B is among them. However, B may not lead to A. & Product positioning (A) depends on consumer evaluation of product attribute (B). Long run profit (A) depends on product positioning (B).\\
\hline
\end{tabular}
\caption{Relations with explanation and examples}
\label{tab:relations}
\end{table}
\fi

\textbf{Knowledge Integration}: For integrating the knowledge in these soft rules, we follow the idea of projecting the learnt predictor function into a rule-regularized subspace \cite{hu-etal-2016-harnessing}. The authors propose a generic way to learn a teacher distribution from a student distribution and a set of rules. Essentially, the teacher ($t(y|x)$) is learned by optimizing the KL-divergence with the student and the constraints imposed by the grounded rules, as follows:  
\begin{equation}
\begin{aligned}
    \min_{t,\Psi \geq 0} & KL(t(y|x) || p_\theta(y|x)) + C \sum_{n,g_n} \psi_{n,g_n}\\
    \textrm{s.t.} \quad & \lambda_n (1- \mathrm{E}_t[r_{n,g_n}(\mathbf{x},y)])\\
    \quad & g_n=1,2,\ldots, G_n; n=1,\ldots, N.
\end{aligned}
\end{equation}
As hard rules evaluate to 1.0, these constraints try to ensure that $\mathrm{E}_t[r_{l,g_l}(\mathbf{x},y)]$ should be as close to 1. Solving the above equation amounts to computing a closed-form solution as given in Equation 4 in \cite{hu-etal-2016-harnessing}, which we reproduce here for convenience: 
\begin{equation}
    t(y|\mathbf{x}) \propto p(y|\mathbf{x})\exp\Big\{- \sum_{n,g_n} C\lambda_n (1 - r_{n,g_n}(\mathbf{x},y)) \Big\}.
    \label{eq:tchr}
\end{equation}
To calculate the second term, we use concepts from MLN inference and T-Norm equations. Primarily, for a predicate $y$ and the input $n_s, n_o$ (i.e., ignoring the sentence information), we assume truth-value of $y(n_s,n_o)$ to be 1 and calculate the value $\lambda_n (1 - r_{n,g_n}(\mathbf{x},y))$ for each grounding of each rule. Essentially, this provides an estimate of number of grounded rules satisfied by the query $y(n_s, n_o)$. Here, the truth value of a grounded rule is computed using Lukasiewicz's T-norm equations. This is a sharp departure from the way this equation is computed in practice by \cite{hu-etal-2016-harnessing}\footnote{As discussed in \cite{krishna2018revisiting}, the mathematical equations do not fully match the code released by authors of \cite{hu-etal-2016-harnessing}}. Overall, we change slightly the DualRE learning algorithm to Algorithm \ref{algo:triple}. Equation \ref{eq:tchr} is computed using the current labeled data $L$ and the set of weighted rules.

\section{Experiments and Results}

To evaluate the pipeline for relation prediction we use an annotated data set. In creating this ground truth from a well-regarded marketing text corpus, out of a total 1748 candidate triples in 231 sentences,  415 triples are annotated by hand. The annotation is done by a marketing expert with more than two decades of consulting and managerial-teaching experience in marketing in the US. In doing this annotation, the expert is provided with relations from ConceptNet. The relation semantics are altered to fit the needs of the domain. A total of 19 relations are used (18 in Table \ref{tab:relations} and one for no relations). Given this annotated dataset, at first we extract the set of features such as tokenized words, parts-of-speech tags, subject and object position indicators for each labeled and unlabeled relation-mentions. In a difference with the DualRE implementation we ignore the object and subject types (and NER tags) as the concepts in our MKG are not named enitites and there is no well-defined ontology to the best of our knowledge.  We use 53 triplets each for validation and test set, and the rest of annotated and unlabeled data are used to create the train set. We use the annotated part as train ($L$) and unannotated part as raw  ($U$)
%\textcolor{red}{what does 'raw' mean?}
according to script in \cite{lin2019learning}\footnote{Code: \url{https://github.com/INK-USC/DualRE}}. For the baseline DualRE, we run their DualRE-pointwise variant. For the rule-regularized version, we run the MLN weight learning algorithm a priori and then provide weighted rules (in Table 2) as inputs to the Algorithm \ref{algo:triple}.
We use the similar EM-based algorithm and run for 10 iterations. 
%\textcolor{red}{why close to?}
We report the final precision, recall and F1 scores for the validation and test set in Table \ref{tab:res1}. 

\begin{table}[!htpb]
\resizebox{0.48\textwidth}{!}{%
\begin{tabular}{|l|l|l|l|l|l|l|}
\hline
             & \multicolumn{3}{c|}{dev}                      & \multicolumn{3}{c|}{test}                     \\ \hline
             & P             & R             & F1            & P             & R             & F1            \\ \hline
DualRE       & \textbf{96.4} & 50.9          & 66.6          & \textbf{59.4} & 41.5          & \textbf{48.8} \\ \hline
DualRE+Rules & 88.9          & \textbf{60.4} & \textbf{71.9} & 28.6          & \textbf{56.6} & 37.9          \\ \hline
\end{tabular}%
}
\caption{Results on the Segmentation chapter. We report precision, recall and F1 scores for both validation and test set.}
\label{tab:res1}
\end{table}

\subsection{Ablation Study}
\begin{table}[!htpb]
\resizebox{0.48\textwidth}{!}{%
\begin{tabular}{|l|l|l|l|l|l|l|}
\hline
             & \multicolumn{3}{c|}{dev}             & \multicolumn{3}{c|}{test}   \\ \hline
             & P    & R             & F1            & P    & R             & F1   \\ \hline
DualRE+Rules & 88.9 & 60.4 & 71.9 & 28.6 & 56.6 & 37.9 \\ \hline
DualRE+R/\{Rj\}  & 88.9 & 60.4 & 71.9 & 28.6 & 56.6 & 37.9 \\ \hline
DualRE+R/\{R2,4,6\}  &  91.17    &   58.5            &   71.3            & 31.4     &     50.9          &  38.8    \\
\hline
DualRE+R5  &  92.0    &   59.6            &   72.3            & 32.5     &     50.5          &  39.54    \\\hline
\end{tabular}%
}
\caption{Ablation study to see the effect of removing each of the rules from the set. }
\label{tab:res2}

\end{table}

% Please add the following required packages to your document preamble:
% \usepackage{multirow}
\begin{table}[!htpb]
\begin{tabular}{|l|l|l|l|l|}
\hline
\multicolumn{2}{|l|}{}               & \multicolumn{1}{c|}{P} & \multicolumn{1}{c|}{R} & \multicolumn{1}{c|}{F1} \\ \hline
\multirow{2}{*}{Affects}         & B & 75                     & 100                    & 85.71                   \\ \cline{2-5} 
                                 & R & 50                     & 100                    & 66.7                    \\ \hline
\multirow{2}{*}{DependsOn}       & B & 54.55                  & 46.15                  & 50.00                   \\ \cline{2-5} 
                                 & R & 44.4                   & 61.5                   & 51.6                    \\ \hline
\multirow{2}{*}{LeadsTo}         & B & 33.33                  & 37.5                   & 35.39                   \\ \cline{2-5} 
                                 & R & 11.3                   & 75                     & 19.7                    \\ \hline
\multirow{2}{*}{MotivatedByGoal} & B & 50                     & 33.3                   & 40.0                    \\ \cline{2-5} 
                                 & R & 25                     & 33.3                   & 28.5                    \\ \hline
\multirow{2}{*}{PartOf}          & B & 90.0                   & 60                     & 72                      \\ \cline{2-5} 
                                 & R & 50                     & 80                     & 61.5                    \\ \hline
\end{tabular}
\caption{DualRE baseline (B) and rule-regularized (R) results for the relations in the test set.}
\label{tab:res3}
\end{table}
One of the contributions of this work is to learn the weights of rules using MLN and integrate this knowledge for improving the accuracy in our relation extraction task. So, as an ablation study, we  experiment with removing each rule and observing the impact on overall scores. The scores are reported in Table \ref{tab:res2}. While we observe that the final scores after removing individual rules do not differ significantly, removing subset of rules makes the end-to-end difference in precision and recall more prominent. We observe, that as we decrease the number of rules precision increases and recall value decreases. In fact, as the set of rules shrinks, we choose to be less restricted in terms of selecting new samples in $L_U$. For convenience, we also show per-relation statistics in the test set in Table \ref{tab:res3}. As our test set is relatively small (because of the limited annotations), most of the other relations occur at most twice and hence we omit them from the table.

%\section{Case Study}

\section{Discussion and Conclusion}
For human managers, marketing decision making is often a complex combination of years of experience in the field, knowledge from text and case studies, and insights from current data. Current technologies provide a peek into utilizing the massive amount of analytics data often available to corporations, but interpreting the data without the lens of knowledge can often send incorrect signals. We intend to bridge the gap by creating a marketing knowledge graph by capturing the knowledge in marketing text. In doing so, the dearth of annotations invokes a well-known, although less-addressed, challenge of predicting relations in a semi-supervised setting. We investigate the effects of integrating hand-coded rules with learned weights as (global) constraints in a semi-supervised relation prediction method and observe improvements. We observe that while trying to learn the rules from a small set of annotated triplets using MLN, the closed world assumption forces the learner to learn only unary clauses. Our current choice of rule integration method leads us to believe that removing a single rule does not affect the results much (and often not at all). Even though, adding rule-based constraints seem to be the intuitive way of integrating prior knowledge in the prediction formulation, final results are not always conclusive. These results yearn for future research in these directions.

%References and End of Paper
%These lines must be placed at the end of your paper

\bibliographystyle{aaai}
\bibliography{references}

\newpage
\appendix
\section{Complete Set of Relations Used}
Here, in table \ref{tab:relations2} we provide the complete set of relations used. Although, it is hard problem to completeness of such relations, to the best of our knowledge the following set of binary relations sufficed to represent the knowledge in the Segmentation chapter in a well-renowned and recent Marketing textbook.

\begin{table*}[!htb]
\begin{tabular}{|p{2.3cm}|p{7.9cm}|p{6.8cm}|}
\hline
Relations & Explanation & Example(s)\\
\hline
LeadsTo & A results in occurrence of B. The occurrence can be through other states, not necessarily direct. & Homogeneous preference (among consumers) (A) leads to competition (more competition) (B)\\
\hline
UsedBy & Usage of A by B for achieving some end state. Applies to both companies and consumers. & Product dimension (A) is used by consumer (B) to make a choice.\\
\hline
ImportantTo & A is a quality or characteristic that is salient to / for B. For a marketer, it is valuable to highlight some characteristics as particularly important, more than merely identifying them as a characteristic. & New dimension of product (A) is important to consumers (B). New dimension of product (A) is important for product positioning (B) by marketer.\\
\hline
DependsOn & A depends on B; however, is different from the inverse of LeadsTo. In marketing, A can depend upon several factors, B is among them. However, B may not lead to A. & Product positioning (A) depends on consumer evaluation of product attribute (B). Long run profit (A) depends on product positioning (B).\\
\hline
Affects & A can have impact on B, does not mean it will have an impact [inverse - AffectedBy]. & Government regulation (A) affects product life cycle (B). Style and fashion affects product life cycle. Political influence affects govenrment regulation.\\
\hline
Enables & A can facilitate the occurrence of B [inverse - EnabledBy]. & Good positioning is enabled by strong advertising claims. Perception and choice consumers form are enabled by product attribute.\\
\hline
% Below are ConceptNet5 relations with modified explanation &  &  \\
\hline
PartOf & A is a characteristic, which marketer associates with B. & Demographics (A) is a part of consumer (B). Price sensitivity (A) is a part of consumer preference (B).\\
\hline
HasPrerequisite & A is achieved by a marketer by performing B. & To determine competitive structure (A) a prerequisite is to understand preference difference [among consumers](B).\\
\hline
MotivatedByGoal & Goal of achieving B motivates marketer to perform A. & Price increase (A) is motivated by goal of pursuing higher profit (B).\\
\hline
HasFirstSubevent & A can start to happen when B starts to occur. & For diffusion of innovation (A) to occur the first subevent of adopt[ing] new product [by consumers](B) is necessary.\\
\hline
CapableOf & Relative to ConceptNet5, CapableOf expanded from "Something that A can typically do is B" to Something that A can do is B, reducing focus on typicality. & Attitude [of consumers](A) is capable of identifying segmentation opportunities for marketer. Attitude is capable of differentiating behaviors [of consumers](B) by marketer.\\
\hline
HasA & A possesses certain traits B. A may not possess always. & Company (A) has strong patents (B). Consumer (A) has a higher price elasticity (B).\\
\hline
HasProperty & B is a property that characterises A; A possesses this property always. & Product (A) has the property that it requires distribution (B).\\
\hline
Synonym & A and B are often considered similar in what they convey. & Attitude segmentation (A) is synonymous with psychographics (B).\\
\hline
UsedFor & Purpose of A is to achieve B. & Perceptual map (A) is used by marketer to identify gaps (B) in marketplace.\\
\hline
CausesDesire & If a marketer experiences A, then it is likely the marketer wants to achieve B. & Shared production cost (A) among different products causes a desire to offer product line across segments (B).\\
\hline
RelatedTo & As in ConceptNet5, interpreted as a general relation. In marketing, many relations take this form, since pin pointing directionality is very difficult, without considering many other factors of context ad environment. & Maturity stage of product (A) in life cycle is related to product's ease of use by consumers (B).\\
\hline
Causes & A can cause B; although not always. In marketing, causal-relations are soft in scope, that is, does not mean A implies B [inverse - CausedBy]. & Good positioning (A) of a product causes high trial rate (B) of the product.\\ 
\hline
\end{tabular}
\caption{Illustrative relations with explanation and examples. The top five relations are new, while others come from ConceptNet5. }
\label{tab:relations2}
\end{table*}

\end{document}